\pdfoutput=1
\documentclass[letterpaper, 10 pt, conference]{ieeeconf}  

\IEEEoverridecommandlockouts

\usepackage{cite}
\usepackage{amsmath,amssymb,amsfonts}
\usepackage{enumerate}
\usepackage{algorithm}
\usepackage{algorithmicx}
\usepackage{graphicx}
\usepackage{textcomp}
\usepackage{xcolor}
\usepackage{float}
\usepackage{stfloats}
\usepackage{subfigure}
\usepackage{booktabs}
\usepackage{siunitx}
\usepackage[noend]{algpseudocode}  
\usepackage{makecell}
\usepackage{hyperref}
\hypersetup{colorlinks = true, 
	linkcolor = black, 
	urlcolor = blue,
	citecolor = blue} 

\def\BibTeX{{\rm B\kern-.05em{\sc i\kern-.025em b}\kern-.08em
		T\kern-.1667em\lower.7ex\hbox{E}\kern-.125emX}}

\title{\LARGE \bf
HybridFusion: LiDAR and Vision Cross-Source Point Cloud Fusion
}

\author{Yu Wang$^{1}$, Shuhui Bu$^{1}$*, Lin Chen$^{1}$, Yifei Dong$^{1}$, Kun Li$^{1}$, Xuefeng Cao$^{2}$, Ke Li$^{2}$
\thanks{*Corresponding author (bushuhui@nwpu.edu.cn)}
\thanks{$^{1}$School of Aeronautics, Northwestern Polytechnical University}
\thanks{$^{2}$PLA Strategic Support Force Information Engineering University}
}

\begin{document}

\maketitle
\thispagestyle{empty}
\pagestyle{empty}

\begin{abstract}

Recently, cross-source point cloud registration from different sensors has become a significant research focus. However, traditional methods confront challenges due to the varying density and structure of cross-source point clouds. 
In order to solve these problems, we propose a cross-source point cloud fusion algorithm called HybridFusion. It can register cross-source dense point clouds from different viewing angle in outdoor large scenes.
The entire registration process is a coarse-to-fine procedure. First, the point cloud is divided into small patches, and a matching patch set is selected based on global descriptors and spatial distribution, which constitutes the coarse matching process. To achieve fine matching, 2D registration is performed by extracting 2D boundary points from patches, followed by 3D adjustment. Finally, the results of multiple patch pose estimates are clustered and fused to determine the final pose.
The proposed approach is evaluated comprehensively through qualitative and quantitative experiments. In order to compare the robustness of cross-source point cloud registration, the proposed method and generalized iterative closest point method are compared. Furthermore, a metric for describing the degree of point cloud filling is proposed. The experimental results demonstrate that our approach achieves state-of-the-art performance in cross-source point cloud registration.

\end{abstract}

\section{Introduction}

With the rapid development of 3D data acquisition technology, numerous methods have emerged that can efficiently and accurately reconstruct 3D data, such as Structure from Motion (SfM), real-time dense 3D reconstruction based on monocular or depth camera vision, and LiDAR mapping.
Although a single sensor may use sequential acquisition to create a comprehensive point cloud, the limited acquisition viewing angle may not cover all target surfaces, resulting in inevitable loss of some content.

For example, DenseFusion \cite{chen2020densefusion} uses the high-altitude UAV viewing angle to conduct real-time 3D dense reconstruction, but only the surface above the objects can be reconstructed, which is difficult to completely describe 3D information of all object surface. 
Moreover, the use of multi-view scanning with a single sensor not only incurs significant time cost but also present a challenge to ensure the stability of the mapping algorithm due to potential variations in scene structures.
Therefore, it is especially important to use multi-view and multi-sensor to collect data and then conduct cross-source point cloud fusion. However, cross-source point cloud fusion is a challenging task due to various difficulties, including differences in density distribution and missing data.

\begin{figure}[tb]
	\centering
	\includegraphics[width=1\linewidth]{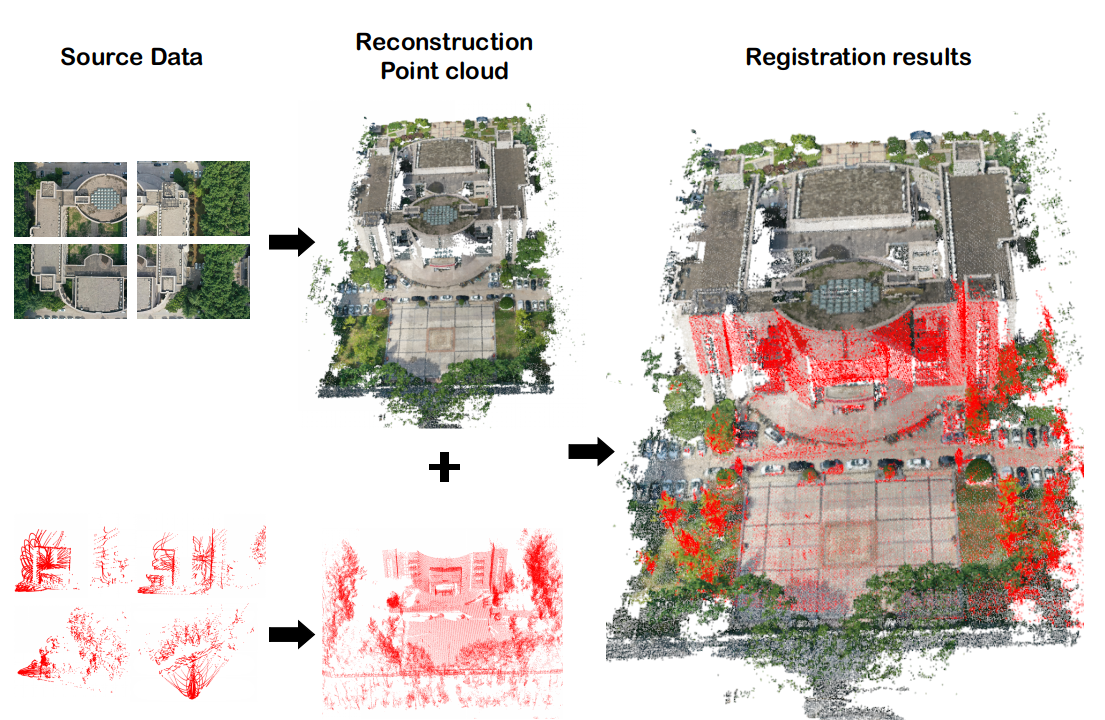}
	\caption{Our method can register cross-source point clouds from different viewing angle. Left part is the dense point cloud generated by vision and LiDAR, and right part is the result of registration.}
	\label{fig:example}
\end{figure}

There are many attempts to solve cross-source point cloud registration. Cross-source graph matching (CSGM) \cite{huang2017systematic} transforms the registration problem into a graph matching problem, subsequently, local Iterative Closest Point (ICP) refinement is conducted, leveraging the global graph matching results. Mellado \textit{et al.} \cite{mellado2015relative} employ a combination of global alignment and local refinement techniques, namely Random Sample Consensus (RANSAC) and Iterative Closest Point (ICP) algorithms, to register point clouds from different sources. Huang \textit{et al.} \cite{huang2019fast} propose a method that consists of two components, weak area affinity and pixel thinning, to maintain the global and local information of 3D point cloud, but this is only support small scenes with high confidence.
Point cloud registration often assumes strong structural consistency, but actual scenes may not have such ideal similarity. Most cross-source point clouds have different densities, noise models and part missing, making it difficult to obtain satisfactory results through conventional point cloud registration methods.

As shown in Fig. \ref{fig:example}, it can be seen that cross-source point clouds often contain missing points, which can present a significant challenge for direct registration. Variations in point density between point clouds can complicate registration, especially when the point clouds are incomplete. To overcome these challenges, the fusion of heterogeneous point clouds can be divided into two stages, resulting in increased robustness and precision.
In the first stage, the point clouds are divided into small point cloud patches in advance to extract effective information, and then the matching candidate sets are obtained by using the global descriptor and spatial distribution of the point cloud patches.
The second stage involves two steps of registration using the Normal Distribution Transform (NDT) algorithm. Firstly, the boundary points in patches projected onto the 2D plane are used for registration, and then 3D fine registration is carried out on this basis, the final pose is obtained by fusing the multiple patch pose. This approach is particularly useful when dealing with missing points in different regions, which cause ICP-based registration to fail.
In summary, contributions of this work are summarized as follows:
\begin{itemize}
	
	\item A multi-view cross-source point cloud fusion algorithm is proposed, which can quickly build a complete 3D point cloud.
	\item In order to overcome the difference in the spatial distribution of point clouds from different viewings, a point cloud registration method based on the joint optimization of global descriptors and 2D boundary is proposed. This method can effectively fuse cross-source point clouds.
	\item A cross-view fusion  dataset\footnote{github.com/npupilab/cvf-dataset}, including multiple simulation and real-world datasets, is established, which is composed of over-view and street-view data. 
	
\end{itemize}

\begin{figure}[t]
	\centering
	\includegraphics[width=1\linewidth]{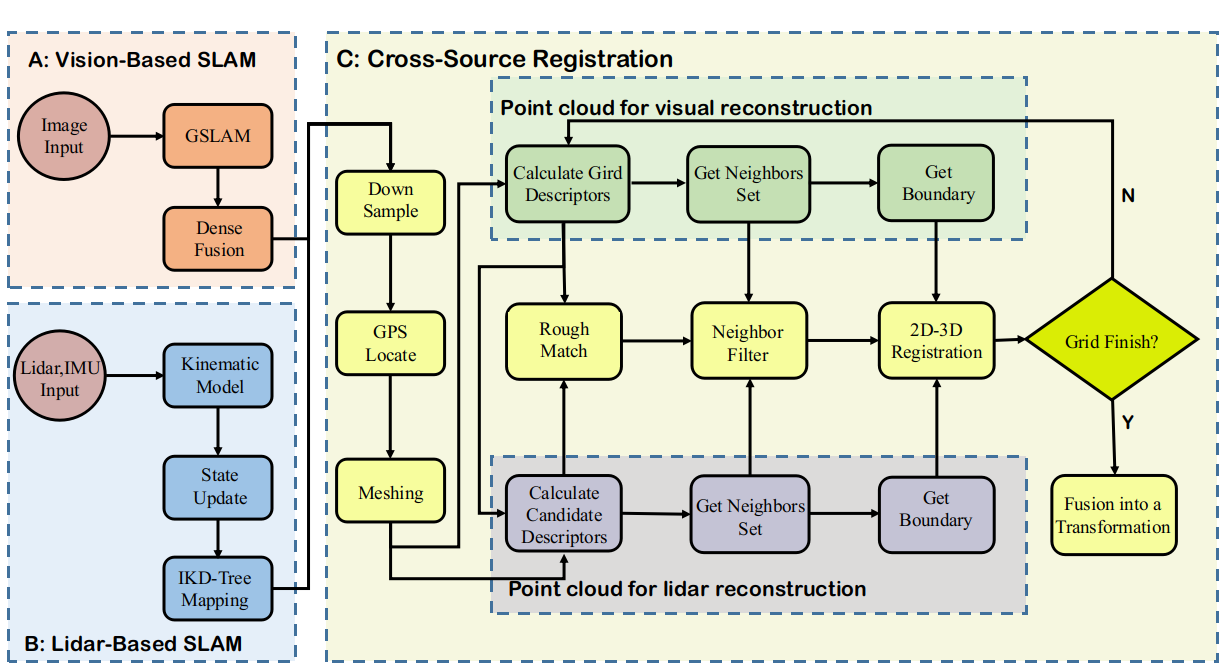}
	\caption{System overview of Hybrid Fusion}
	\label{fig:pipline}
\end{figure}


\section{Related Works}

Our approach relys on three research field: monocular camera dense reconstruction, LiDAR mapping and point cloud registration. Some related works are summarized in follows.

\subsection{Monocular Dense Reconstruction}
Monocular dense reconstruction from image sequences has been well-studied research field in recent years.
The dense matching algorithm used by SfM also has extensive updates. The earliest algorithms, such as the least square matching algorithm \cite{ackermann1984digital,gruen1985adaptive}, have evolved into more advanced methods like patch match stereo (MVS) method. Recent advancements include the semi global matching (SGM) \cite{hirschmuller2007stereo} and patch based MVS \cite{furukawa2010patch}. To solve the depth map reconstruction stereo matching problem, typically, matching cost calculation and aggregation, as well as disparity calculation and refinement, are performed using local, global, or semi-global energy minimization techniques.

Another important research interest is real-time dense reconstruction. Bloesch \textit{et al.} proposed CodeSLAM \cite{bloesch2018codeslam}, which can obtain a compact implicit representation of a dense depth map using only a few parameters. 
Chen \textit{et al.} proposed DenseFusion \cite{chen2020densefusion}, which completed 3D reconstruction by building virtual stereo pairs from selected key frames to generate dense 3D point clouds after trimming.

\subsection{Lidar-based SLAM}

The most important method of LiDAR-based simultaneous localization and mapping (SLAM) system is rely on sparse features \cite{zhang2014loam,moosmann2011velodyne}, such as line features and surface features. For the sparse features, most LiDAR-base SLAM developed from LiDAR odometry and mapping (LOAM) \cite{zhang2014loam}. It is composed of three main modules: feature extraction, odometry, and mapping. To reduce the computation load, local smoothness is used to extract features from each frame of LiDAR scanning. The odometry module matches the extracted features from two consecutive frames to obtain a rough but real-time LiDAR odometry. Finally, the 3D points from the LiDAR scan are registered to the global map.
Subsequent LiDAR-based SLAM works keep a framework similar to LOAM. For example, 
Lego-LOAM \cite{shan2018lego} introduces ground point segmentation to reduce the computation and use a closed-loop module to reduce scene drift. LIOM \cite{ye2019tightly} proposes a tightly-coupled LiDAR inertial fusion method where the IMU pre-integrations are added into the odometry. FAST-LIO2 \cite{xu2022fast} uses tightly coupled iterative Kalman filtering to directly register the original LiDAR points to the map without extracting features. To maintain the map, it employs an incremental k-d tree structure called the Incremental Kalman-filtered k-d tree (IKD-tree).

\subsection{Point Cloud Registration}

Our method mainly involves two types of point cloud registration: feature-based and NDT-based methods.
Featured-based methods extract the feature from point cloud, and transforms the point cloud registration 3D place into feature space. Feature-based registration has two interests, the former extract features from the fields near a feature point of the point cloud, such as binary shape context (BSC) \cite{dong2018hierarchical}, Fast Point Feature Histograms (FPFH) \cite{rusu2009fast}, and rotational projection statistics (RoPS) \cite{guo2013rotational}.
However, these methods often encounter difficulties in cross-source point cloud registration due to differences in point cloud density and scanning range across different sources. The latter is a feature extraction method applied to the entire point cloud, describing its the overall characteristics, such as Ensemble of Shape Functions (ESF) \cite{wohlkinger2011ensemble}, Globally Aligned Spatial Distribution (GASD) \cite{do2016efficient}, and so on. 
The global descriptor is better than those feature-based points for cross-source point cloud registration, which pays more attention to the features of the whole point cloud rather than the local region.

The Normal Distribution Transform (NDT) method compares the normal distribution of points in a voxel grid between two point clouds to find the best match. It uses standard optimization techniques to minimize the difference between the distributions. 
Das \textit{et al.} \cite{das2012scan} proposed a multi-scale k-means normal distribution transformation (MSKM-NDT), which addresses the issue of cost discontinuity in point cloud registration by dividing the data into clusters using k-means clustering and optimizing them at various scales. Lu \textit{et al.} \cite{jun2015point} proposed a nondestructive testing algorithm of variable size voxels, which improved the accuracy of the algorithm. 
NDT registration is to register point clouds by comparing the similarity of grids and nonlinear optimization, so it has certain robustness to cross-source point clouds. In the subsequent algorithm of this paper, the NDT algorithm is used for registration.


\section{Methodology}

The overall processing pipeline is shown in the Fig. \ref{fig:pipline}. It takes a sequence of images, LiDAR scans and IMU data as input, while high quality fused point clouds are generated.

The \textit{module-A} takes charge of visual dense reconstruction. In this module, UAV is used to take aerial photos, whose position is estimated with the SLAM plug-in using the GSLAM framework. At the same time, in order to fix the scale of the visual reconstruction point cloud, the GNSS information is integrated into the optimization process. Finally, the DenseFusion method \cite{chen2020densefusion} is adopted to densify the point cloud.

The \textit{module-B} adopts FAST-LIO2 algorithm \cite{xu2022fast} to generates dense point cloud. It uses an efficient tightly-coupled iterated Kalman filter, and directly registers the raw points to the map without extracting features. Then, IKD-tree is used to dynamically maintain the map, which has better search and access speed than traditional methods.

The \textit{module-C} performs cross-source point cloud registration. Firstly, mapping LiDAR-based clouds into vision-based ones using GNSS position information. In addition, the approximate range obtained by the visual SLAM can also be used to replace GNSS information. Due to the presence of large GNSS error, only a coarse registration range can be determined. Point clouds are divided into small patches, and candidate patches are selected based on similar information. However, this alone cannot solve the problem of cross-source point cloud registration. To overcome limitations, a novel cross-view point cloud registration method is proposed. This method uses the boundary points of 2D projection for rough registration, and then performs 3D fine registration. The registered position of all patches is fused into a final position, resulting in precise cross-source point cloud registration with different viewing angles.

\subsection{Dense Reconstruction}

Real-time monocular dense reconstruction is the initial phase. It consists of two parts: the pose estimation algorithm and dense reconstruction.

\subsubsection{Pose Estimation Algorithm}

For each frame of image, SiftGPU \cite{wu2011siftgpu} is used to extract feature points and corresponding descriptors.
In order to acquire the georeferenced position, the drone's pose estimation integrates both visual and GNSS-based positions. The Bundle Adjustment (BA) process and optimize two loss functions: re-projection error $\boldsymbol{e}_ r$ and GNSS error $\boldsymbol{e}_g $. The overall error function is defined as follows:
\begin{equation}
\min f(x)=\frac{1}{2}\left[\sum_{i=0}^n \boldsymbol{e}_{r_i}^{\top} W_r \boldsymbol{e}_{r_i}+\boldsymbol{e}_g^{\top} W_g \boldsymbol{e}_g\right],
\end{equation}
where $\boldsymbol{e}_{r_i}$ is re-projection error for 3D points, $\boldsymbol{e}_{g}$ represents the scale error of both visual and GNSS, $W_r$, $W_g$ are the information matrix of re-projection error and GNSS error. $\boldsymbol{e}_{r}$ is defined as follows:
\begin{equation}
\boldsymbol{e}_{r}=\boldsymbol{x}_i^{\prime}-\frac{1}{s_i} \boldsymbol{K} \exp \left(\boldsymbol{\xi}^{\wedge}\right) \boldsymbol{P}_i,
\end{equation}
where $\boldsymbol{x_{i}^{'}}$ stands for the position of corresponding key point in current frame. $\boldsymbol{K}\in \mathbb{R}^{3 \times 3}$ is the camera internal parameter matrix, $ \boldsymbol {\xi} \in \mathfrak{se}(3)$ represents the camera position and pose represented by Lie algebra, and $\boldsymbol {P_i} \in \mathbb {R}^{3}$ represents the map point position.

GNSS Error $\boldsymbol{e}_g$ is defined as follows:
\begin{equation}
\boldsymbol{e}_{g}=\boldsymbol{t}_{GNSS}-\boldsymbol{t}_{SLAM},
\end{equation}
where $\boldsymbol{t}_{GNSS}$ represents the displacement vector of GNSS between two frames, $\boldsymbol{t}_{SLAM}$ represents the displacement vector of SLAM.

\subsubsection{Densification}

To ensure point cloud accuracy, dense reconstruction uses larger baseline images derived from adjacent time frames as stereo image pairs. Bouguet's rectification aligns stereo images with parallel optical axes, and CUDA-accelerated SGM \cite{hirschmuller2007stereo} performs efficient stereo matching.

To minimize depth inaccuracies, we utilized a combination of depth consistency verification and photometric consistency verification.

\subsection{Lidar-based SLAM}

In order to obtain high quality LiDAR point clouds, FAST-LIO2 \cite{xu2022fast} is adopted, which primarily consists of two modules: the motion model of tightly coupled Kalman filter and the mapping module.

\subsubsection{Motion Model}

The Kalman filter-based motion model is employed for efficient processing, which performs forward propagation of IMU measurements and iterative updating of each LiDAR scan. The forward propagation estimates the current LiDAR position by integrating the IMU measurements when they arrive. Specifically, it transfers the current LiDAR state and covariance when the process noise is small. Based on the IMU forward propagation estimation, the scanned point cloud is projected into the world coordinate system. The point cloud is then aligned with the map points using nonlinear optimization to minimize errors, resulting in a more accurate pose estimation.

\subsubsection{Mapping}

The novel data structure IKD-tree is used to organize the map points for large scale range mapping. To control the size of the map, the system saves a specific length of area around the current location. When the LiDAR scanning area touches the map boundary, the map moves in the direction of the boundary and deletes map points in the opposite direction. All of these operations are based on the IKD-tree data structure.

\begin{figure}[t]
	\centering
	\includegraphics[width=0.8\linewidth]{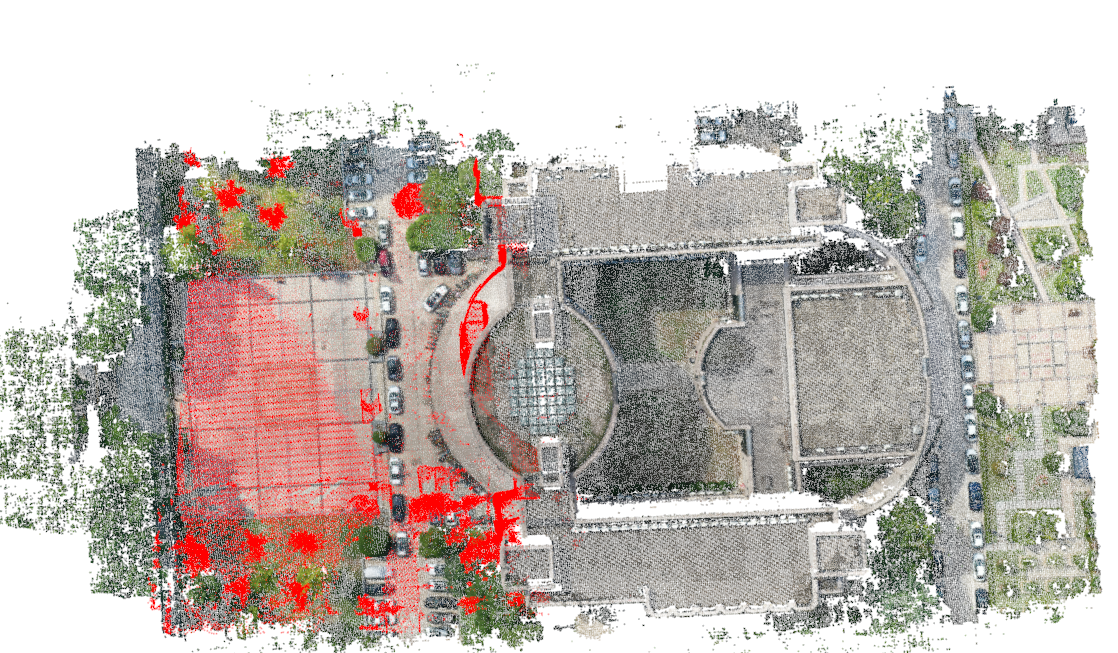}
	\caption{GNSS Registration, red one is LiDAR-based point cloud,others are the vision-based point cloud. There is an obvious mismatch in the cylindrical tower.}
	\label{fig:GPSRegister}
\end{figure}

\subsection{Cross-Source Registration}
 
After generated two point clouds from dense reconstruction and LiDAR-based SLAM, a cross-source point cloud registration is required to generated a high quality point cloud. 
Our algorithm can register point clouds from different viewing angles. The detailed processing flow is described in following sections.

\subsubsection{Pre-processing}

Appropriate pre-processing, which is separated into initial adjustment and point clouds meshing, is required for dense point clouds.
The initial adjustment of point cloud mainly includes: point cloud down sampling and coarse alignment. 
The first step is the point cloud down sampling. The VoxelGrid filter \cite{Rusu_ICRA2011_PCL} is used for down sampling the point cloud, which reduces its density without losing details. 
The second step is coarse alignment which use GNSS constraints. 
However, the accuracy of GNSS is often limited, leading to inaccurate alignment, as shown in Fig. \ref{fig:GPSRegister}. Typically, only the approximate positioning range can be determined.

Point cloud gridding:
For cross-source point clouds, only a few sub-regions are similar due to significant differences in regional feature distributions. To address this issue, the point cloud is segmented into smaller patches for individual registration. Then, the overall registration pose of the point cloud can be estimated based on the registered patches.
To partition the point cloud into patches, the point cloud area is divided into a grid using a specified step length. The step length is set to one tenth of the scene size based on experience. Points are then assigned to different grids based on their positions. 

We define the visual dense reconstructed point cloud as $ \mathbf{G}= \left \{g_1, g_2, g_3 \cdots g_n \right \}$ and the LiDAR reconstructed point cloud as $\mathbf{L}= \left \{l_1, l_2, l_3 \cdots l_m \right \}$. The point cloud patches in $\mathbf{G}$ are defined as $\mathbf{P}_\mathbf{G} = \left\{p_{g_1},p_{g_2}, p_ {g_3} \cdots p_{g_u} \right\}$. In the same manner, point cloud $\mathbf{L}$ is also divided into patches as $\mathbf{P}_\mathbf{L} = \left \{p_{l_1}, p_ {l_2}, p_ {l_3} \cdots p_ {l_w}\right \}$.

\subsubsection{Coarse Match}

Even though point clouds $\mathbf{G}$ and $\mathbf{L}$ lack complete surface data, there still exist some overlapping regions that can be utilized to approximately match their patches.
The proposed method divides this into two steps: selecting point cloud candidates and filtering them.

\paragraph{Candidate Selecting}

The candidate range of the patches has a great impact on the subsequent matching accuracy,  therefore it is necessary to select a good candidate range of patches. 

The first step is to identify significant features in the LiDAR point cloud patches. The selected patches, denoted as $\mathbf{P}^{slt}_\mathbf{L}$, are chosen from $\mathbf{P}_\mathbf{L}$ based on the criteria of having a point count in a patch greater than $\eta$. This threshold is set as insufficient patches do not possess adequate discriminate.

\begin{figure}[t]
	\centering
	\includegraphics[width=0.8\linewidth]{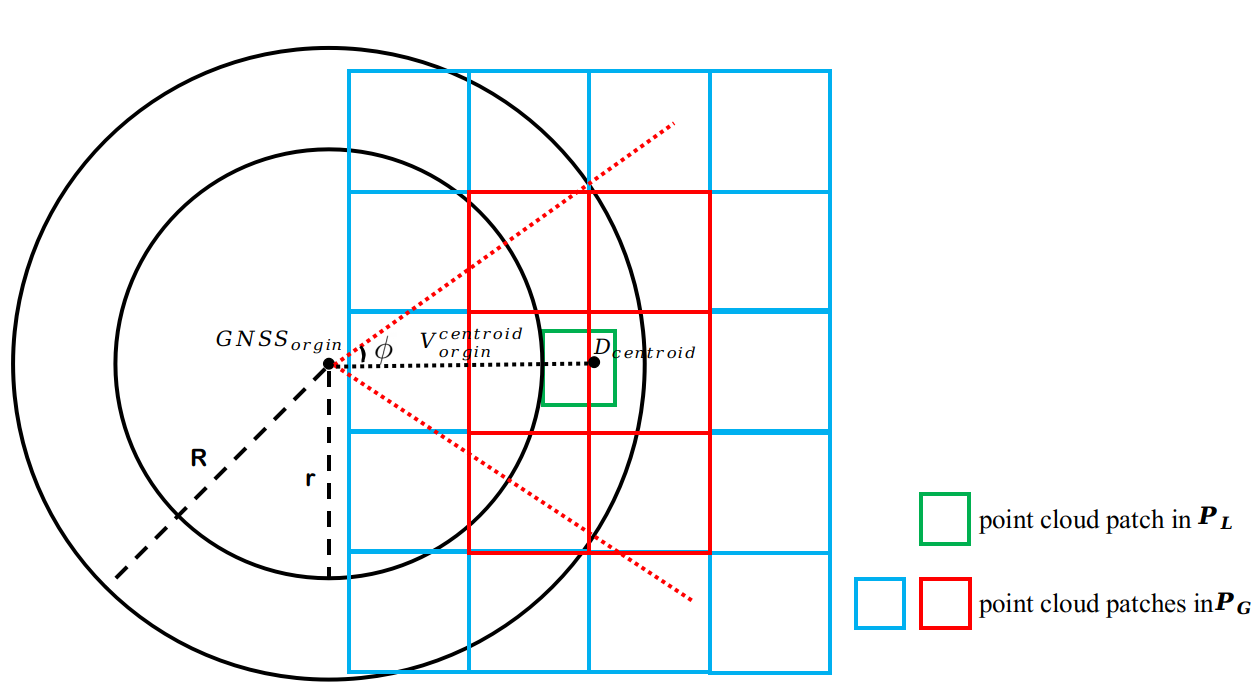}
	\caption{2D demonstration of point cloud selection. Green patch is the point cloud patch in $\mathbf{P}_\mathbf{L}$, blue red patches are in $\mathbf{P}_\mathbf{G}$, red patches mean the candidate patches we select, while blue indicates it is not a candidate. \textbf{$\phi$} is the limited angle for selecting the candidates.}
	\label{fig:candidates}
\end{figure}

The patches in $\mathbf{G}$ are selected based on the position of $\mathbf{P}^{slt}_\mathbf{L}$, as shown in Fig. \ref{fig:candidates}. The origin of $\mathbf{L}$ is mapped to $\mathbf{G}$ through GNSS, resulting in $\mathbf{GNSS}_{orgin}$. For each $p^{slt}_{l_i}$ in $\mathbf{P}^{slt}_\mathbf{L}$, its centroid is defined as $ \mathbf{D}_{centroid}$ and the Euclidean distance $E$ between $\mathbf {GNSS}_{orgin}$ and $\mathbf {D}_{centroid}$ in the $X-Y$ plane is calculated. To determine the candidate point cloud patches range, a circular ring on the $X-Y$ plane is defined with a center at $\mathbf{GNSS}_{orgin}$. The radii of the ring are defined as: 
\begin{eqnarray}
R & = & (1+\lambda)E, \\
r & = & (1-\lambda)E,
\end{eqnarray}
where $\lambda$ is the scaling factor for determining the radius range. $C^R_r$ denotes the circle ring with inner radius of $r$ and outer radius of $R$.  
To find candidate patches, point cloud patches are selected if any point falls in the $C^R_r$ and then point cloud patches with $\theta$ satisfying $-\phi \leq \theta \leq \phi$ are reserved, where $\theta$ is the angle determined by the line connecting the patch centroid and $\mathbf {GNSS}_{orgin}$, as well as the line $\mathbf{V}^{centroid}_{orgin}$, as shown in Fig. \ref{fig:candidates}. The candidate point cloud patches are stored into the set $\mathbf{P}^{cdt}_{\mathbf{G}\rightarrow l_i}$.

\paragraph{Patch Filtering}

The use of descriptors helps to compare each selected LiDAR point cloud patch ($p^{slt}_{l_i}$) with its corresponding candidate patches ($\mathbf{P}^{cdt}_{\mathbf {G}\rightarrow l_i}$). The candidate patches with low similarity are removed to improve the precision of future registration. The ESF640 descriptor is used, but due to its high dimension, ordinary descriptor similarity calculation, such as Euclid distance is computation consuming. Hence, the Pearson correlation coefficient is used instead. 
\begin{equation}
\rho(x, y)=\frac{\sum_{i=1}^n\left(x_i-\bar{x}\right)\left(y_i-\bar{y}\right)}{\sqrt{\sum_{i=1}^n\left(x_i-\bar{x}\right)^2} \sqrt{\sum_{i=1}^n\left(y_i-\bar{y}\right)^2}}.
\end{equation}
Here, $x$ and $y$ represent descriptors. $\bar{x}$ and $\bar{y}$ respectively represent the mean values of $x$ and $y$. It can be seen that the calculation can be optimized by pre-calculating the $(x_i - \bar{x})$ and $(y_i - \bar{y})$, so that the calculation will be greatly accelerated.
If a patch $p^{cdt}_{g_j \rightarrow l_i}$ in $\mathbf{P}^{cdt}_{\mathbf {G}\rightarrow l_i}$ and its corresponding patch $p^{slt}_{l_i}$ do not satisfy the condition $\rho(p^{slt}_{l_i}, p^{cdt}_{g_j \rightarrow l_i})>0.6$, the two descriptors are regarded as irrelevant and the $p^{cdt}_{{g_j\rightarrow l_i}}$ will be remove from $\mathbf{P}^{cdt}_{\mathbf{G}\rightarrow l_i}$.

\subsubsection{Neighbor Filter}

According to the above preliminary filtering, the LiDAR point cloud patches and its similar visual candidate patches have been obtained, but there are still some patches with poor similarity that need to be eliminated. This requires comparing the similarity among the neighbors of the point cloud patches.
First, we need to get the neighbors of each patch in $\mathbf{P}^{slt}_\mathbf{L}$ and its related $\mathbf{P}^{cdt}_{\mathbf {G}\rightarrow l_i}$ sets. If the number of points in its neighbor point cloud patch is greater than a threshold, it will be selected into corresponding neighbor patch set whose descriptor should be calculated.

For each patch $p^{slt}_{l_i}$ in  $\mathbf{P}^{slt}_\mathbf{L}$, and its corresponding candidate set $\mathbf{P}^{cdt}_{\mathbf{G}\rightarrow l_i}$, compare their descriptors within their neighborhoods. Then record its average descriptor similarity value.  
Next, if the value is lower than a certain threshold, the patch $p^{cdt}_{g_j \rightarrow l_i}$ will be removed from the $\mathbf{P}^{cdt}_{\mathbf{G}\rightarrow l_i}$.

Remember that matching in the same direction between neighbor sets is not mandatory (e.g. a patch's upper neighbor does not have to match with another patch's upper neighbor). This is because the point cloud segmentation process uses different criteria, which may result in poor matches if forced to match in the same direction. 
Therefore the matching in the same direction is not mandatory required.

\begin{algorithm}[htbp]
	\scriptsize
	\caption{ \scriptsize 2D-3D Registration} 
	\label{alg:2D-3D_Registration}
	\hspace*{0.02in} {\bf Input:} 
	$p^{slt}_{l_i}$,$p^{cdt}_{g_j \rightarrow l_i}$\\
	\hspace*{0.02in} {\bf Output:}
	$\mathbf{K}$
	\begin{algorithmic}[1]
		\State $\mathbf{P}^{slt+nbr}_\mathbf{L}$ = splicing\_neighbors($p^{slt}_{l_i}$)
		\State $\mathbf{P}^{cdt+nbr}_{\mathbf{G}\rightarrow l_i}$ =  splicing\_neighbors($p^{cdt}_{g_j \rightarrow l_i}$)
		\For{$p^{slt+nbr}_{l_i}$ in $\mathbf{P}^{slt+nbr}_\mathbf{L}$}
		\For{$p^{cdt+nbr}_{g_j\rightarrow l_i}$ in $\mathbf{P}^{cdt+nbr}_{\mathbf{G}\rightarrow l_i}$}
		\State $p^{bdy}_{l_i}$ = project2D\_get\_boundary\_points($p^{slt+nbr}_{l_i}$)
		\State $p^{bdy}_{g_j\rightarrow l_i}$ = project2D\_get\_boundary\_points($p^{cdt+nbr}_{g_j \rightarrow l_i}$)
		\State score\_2d = find\_min\_2D\_score($p^{bdy}_{l_i}$, $p^{bdy}_{g_j\rightarrow l_i}$)
		\If{score\_2d $\leq$ min\_score}
		\State $p^{cdt+nbr}_{g_{min} \rightarrow l_i}$ = find\_min\_2d\_score\_patch(score\_2d)
		\State $\mathbf{T}_{2d}$ = get\_2D\_trans(score\_2d) 
		\State min\_score = score\_2d 
		\EndIf
		\EndFor
		\State score\_3d = find\_min\_3D\_score($p^{slt+nbr}_{l_i}$,$p^{cdt+nbr}_{g_{min} \rightarrow l_i}$)
		\State $\mathbf{T}_{3d}$ = get\_3D\_trans(score\_3d)
		\If{NDT\_3D\_convergence(score\_3d)} 
		\State $\mathbf{T}_{patch}$ = $\mathbf{T}_{3d}*\mathbf{T}_{2d}$
		\State $\mathbf{K}$.push\_back($\mathbf{T}_{patch}$)
		\EndIf
		\EndFor
		\State \Return $\mathbf{K}$
	\end{algorithmic}
\end{algorithm}

\subsubsection{2D-3D Registration}

The matched set for each element of $\mathbf{P}^{slt}_L$ has been roughly found. Point cloud registration will be performed next. The surface of point clouds may be incomplete due to different viewing angles, and ESF descriptor can only help to some extent, but it is not suitable for high accurate registration.
However, for different reconstruction viewing angle, the boundary points of the object are mostly complete, so it can be used to register.

\begin{figure}[t]
	\centering
	\includegraphics[width=1.0\linewidth]{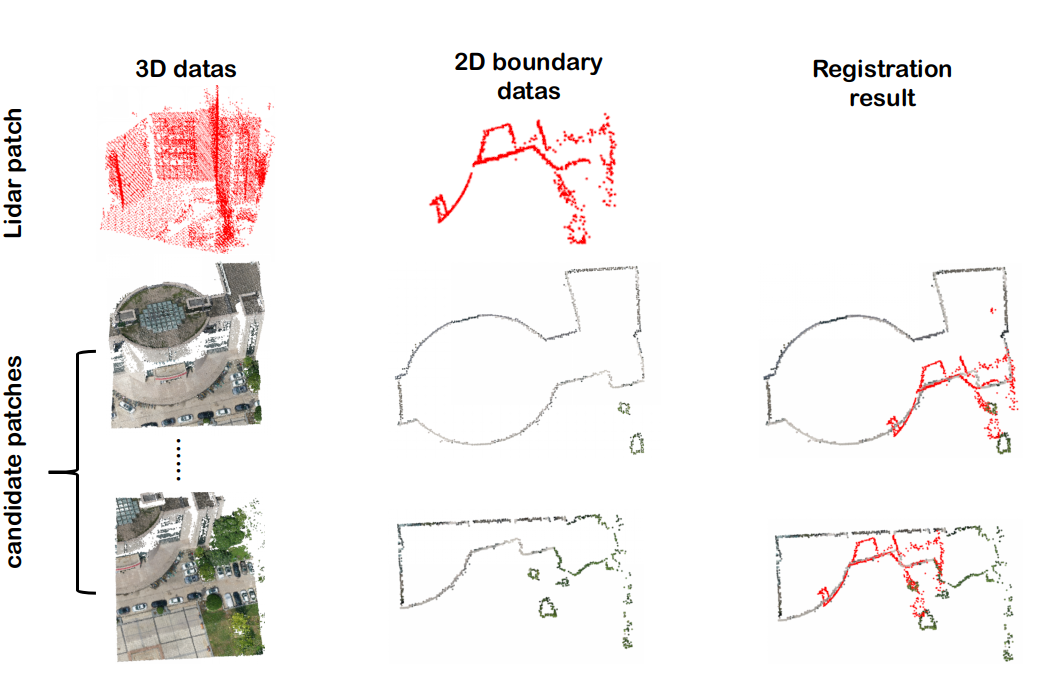}
	\caption{2D boundary points registration. LiDAR patch and its candidate patch set are projected onto the $z$-plane and boundary points are extracted, then registered on this basis, which can overcome the problems of point cloud missing and inconsistent structure to a great extent.}
	\label{fig:2D_boundary}
\end{figure}

Two large point cloud patch sets for registering are denoted as $\mathbf{P}^{slt+nbr}_\mathbf{L}$ and $\mathbf{P}^{cdt+nbr}_{\mathbf{G}\rightarrow l_i}$, which are generated by combining original $\mathbf{P}^{slt}_\mathbf{L}$ and $\mathbf{P}^{cdt}_{\mathbf{G}\rightarrow l_i}$ with their neighbor sets.

In the following processing, each point cloud patches $p^{slt+nbr}_{l_i}$ in $\mathbf{P}^{slt+nbr}_\mathbf{L}$ are used for estimating their own transformation. Boundary points are extracted for each $p^{slt+nbr}_{l_i}$ and its corresponding set $\mathbf{P}^{cdt+nbr}_{\mathbf{G}\rightarrow l_i}$. Points with $d>h$ are extracted from each patch to filter out ground noise, where $d$ is the distance from the point to the ground plane and $h$ is the height threshold value.
Then, the collected points are projected to the $X-Y$ plane, and the boundary points are calculated, through the boundary estimation algorithm \cite{Rusu_ICRA2011_PCL}.
Next, the two-dimensional NDT algorithm is used, $p^{slt+nbr}_{l_i}$ is registered with each element in its corresponding set $\mathbf{P}^{cdt+nbr}_{\mathbf{G}\rightarrow l_i}$. An example of 2D registration is shown in the Fig. \ref{fig:2D_boundary}.

Finally, through the cost of the minimum average distance of points the $p^{slt+nbr}_{l_i}$ with it best matching point cloud patch can be found. In addition, their optimal transformation $\mathbf{T}_{2d}$ is also estimated.

Based on 2D registration results, a 3D registration is conducted through 3D NDT, which adjusts pose in the $z$-axis direction. The registration will be adopted if the matching process can converge, otherwise it will be abandoned.

After 3D registration process, the transformation set $\mathbf{T}_{3d}$ is obtained, the registration pose of this $p^{slt+nbr}_{l_i}$ is defined as $\mathbf{T}_{patch} = \mathbf{T}_{3d} \cdot \mathbf{T}_{2d}$.  
If one pair point cloud patches can estimate the transformation $\mathbf{T}_{patch}$, it is append to the final results set $\mathbf{K} = \mathbf{K} \cup \mathbf{T}_{patch}$.
The detail of 2D-3D registration is depicted in Algorithm \ref{alg:2D-3D_Registration}.

\subsubsection{Post Processing}

After the transformation set is estimated, an overall optimized transformation is fused from the transformation set $\mathbf{K}$. The transformation ($\mathbf{T}$) in $\mathbf{K}$ is split into $\mathbf{Q}$ (a quaternion representing rotation) and $\mathbf{t}$ (representing translation), which represented in the four-dimensional space ($x$, $y$, $z$, $\theta$), where $x$, $y$, $z$ represent $\mathbf{t}$ and $\theta$ represents the angle of two quaternions. Because each $p^{slt+nbr}_{l_i}$ have its own transformation, cluster according to the following requirements:
\begin{equation}
\begin{aligned}
&\sqrt{(x_i-x_j)^2 + (y_i-y_j)^2 + (z_i-z_j)^2 } < \epsilon, \\  
& \qquad \qquad fabs\{R[\mathbf{Q}_i,\mathbf{Q}_j]\} < \omega,
\end{aligned}
\end{equation}
where, $R[\mathbf{Q}_i, \mathbf{Q}_j]$ represents the angle between $\mathbf{Q}_i$ and $\mathbf{Q}_j$.
If both translation and angle between two different transformation are less than the threshold values $\epsilon$ and $\omega$, they can be grouped into one cluster. After all elements in the $\mathbf{K}$ are processed, the cluster with the largest number of transformations is extracted and its average value is calculated as the optimal transformation, where the average value of $\mathbf{Q}$ is calculated by spherical linear interpolation. The registration can be completed by transforming $\mathbf{L}$ to $\mathbf{G}$. Later, minor adjustments can be adjusted through 3D NDT once again.

\begin{figure*}[htbp]
	\centering
	\includegraphics[width=0.8\linewidth]{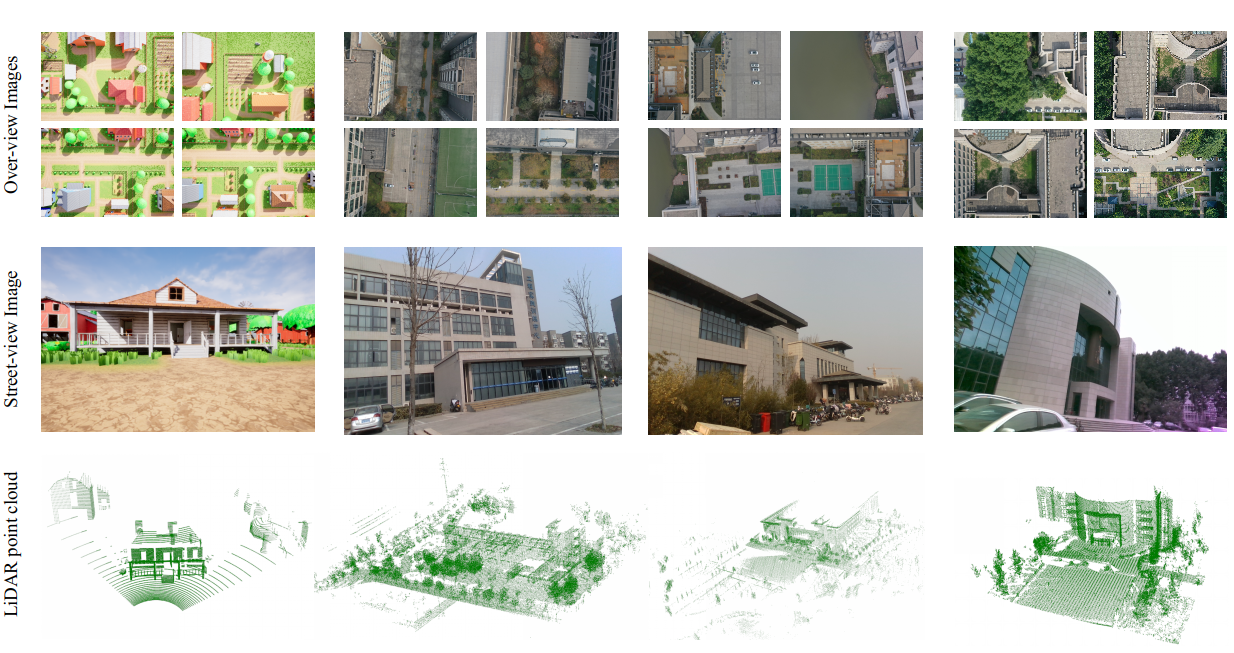}
	\caption{Cross-source dataset. The first row is the over-view image set taken by UAV. The second row is a street-view image example taken by a UGV. The last row is the point cloud generated by the LiDAR. }
	\label{fig:dataset}
\end{figure*}


\section{Experiments}

In order to evaluate the performance of the proposed method, we first establish multiple simulated and real-world datasets, demonstrating the registration results on these datasets.  Then, evaluation metrics are proposed to assess the completeness and alignment accuracy of the point clouds. The results show that our system achieves highly robust and accurate cross-source point cloud registration.

The algorithm is implemented based on C++, and all experiments are tested on a desktop PC equipped with Intel i7-8700 CPU and 16 GB RAM. We run our approach in 64-bit Linux system.

\subsection{Evaluation Metrics}

For evaluating performance of point cloud registration, we propose a novel evaluation metric. First, the result of the point cloud after registration is defined as $\mathbf{O}$. Then, $\mathbf{O}$ and point cloud $\mathbf{G}$ reconstructed from monocular are described by octree. The evaluation metric of point cloud supplement degree is defined as follows:
\begin{equation}
S_{pt}= \frac{ V_\mathbf{O} - V_\mathbf{G}}{V_\mathbf{G}},
\end{equation}
where $S_{pt}$ represents the supplement degree, $V_\mathbf{O}$ and $V_\mathbf{G}$ represent the leaves volume of $\mathbf{O}$ and $\mathbf{G}$ after being wrapped by octree. 

For the accuracy evaluation of registration results, due to the missing of point clouds, the traditional methods cannot be used. However, the point cloud boundary is usually complete, the accuracy is compared by detecting the nearest distance between the boundary points of the point clouds. 
In the evaluation, the average value of nearest distance is obtained as the registration accuracy. 
\begin{figure*}[htbp]
	\centering
	\includegraphics[width=0.9\linewidth]{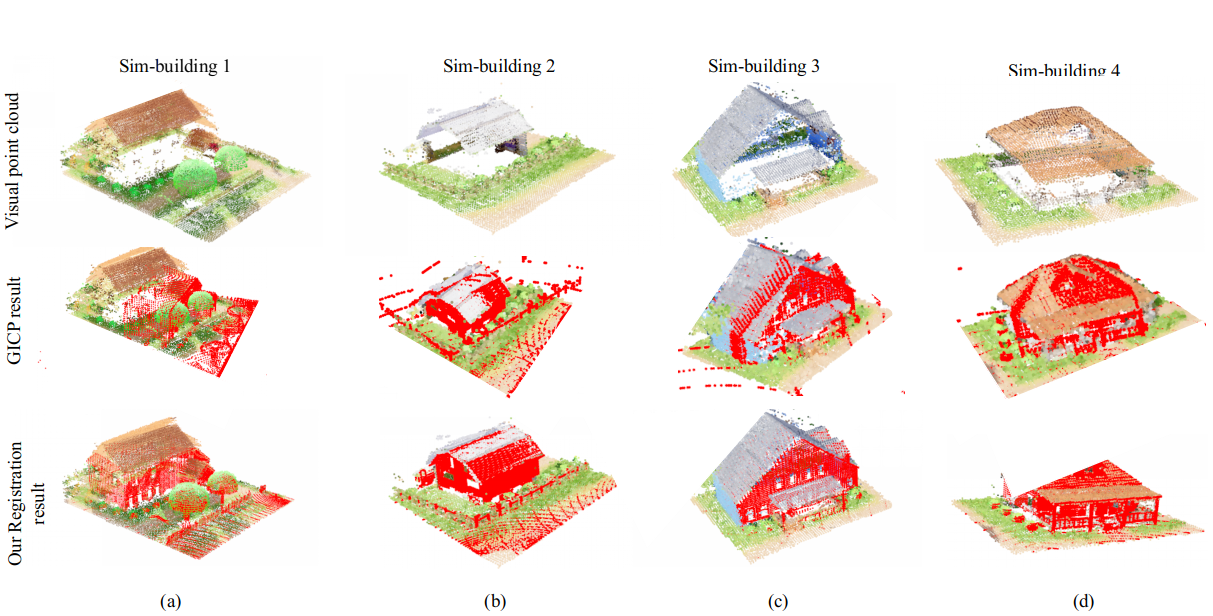}
	\caption{Simulation experiment results. The first row represents the dense point cloud reconstructed from an overhead view, the second row represents the LiDAR fusion result obtained using the GICP method, and the third row represents the LiDAR fusion result obtained using our method.}
	\label{fig:sim_ex}
\end{figure*}

\begin{figure*}[htbp]
	\centering
	\includegraphics[width=0.9\linewidth]{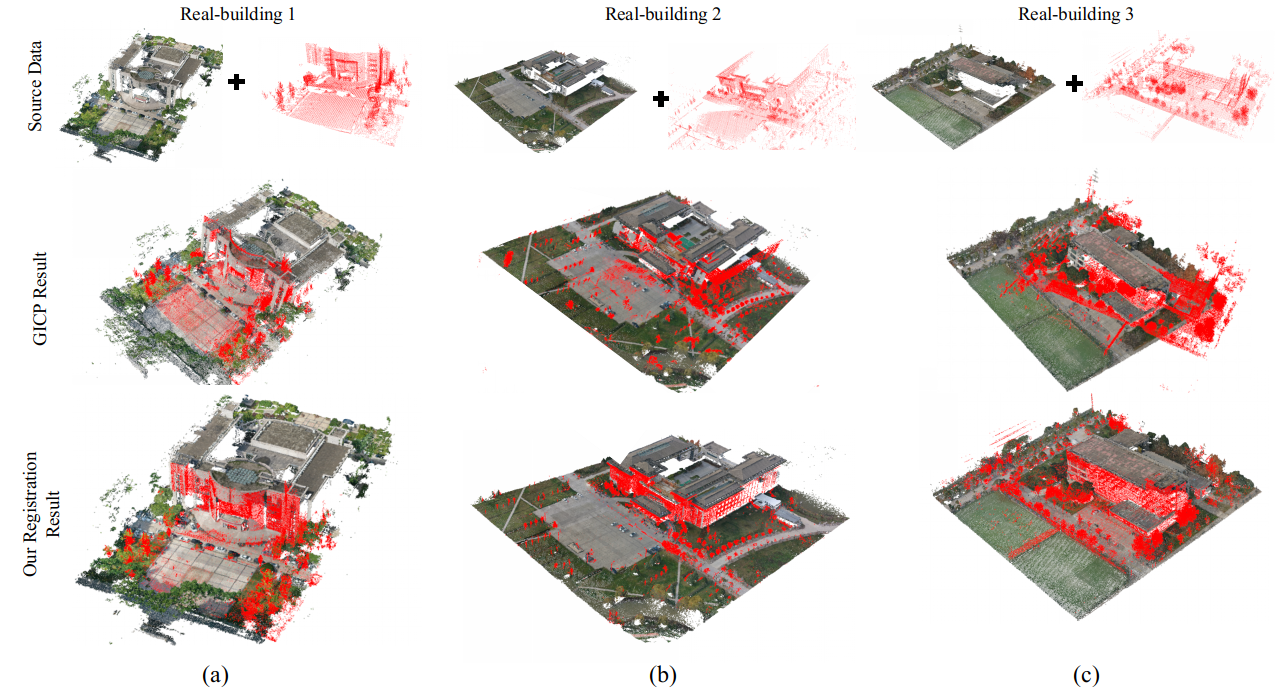}
	\caption{Real world experiment results. The first row represents the dense point cloud reconstructed from an overhead view and the LiDAR point cloud. The second row represents GICP method result. The third row represents our method result. }
	\label{fig:real_ex}
\end{figure*}

\begin{table*}[htbp]
	\caption{Point Cloud Result Evaluation Metrics}
	\label{table:evaluation_metrics}
	\centering
	\begin{tabular}{|l|S|S|S|S|S|}
		\hline
		Scene                   &{\makecell{Visual point cloud volume ($m^3$)}}  &{\makecell{Result volume ($m^3$)}} &{\makecell{Volume growth ratio}}  &{\makecell{GICP Accuracy ($m$)}} &{Our Accuracy ($m$)}\\ \hline
		\textit{sim-building1}  & 217.49    			& 287.57  				& 32.21\%    &2.54          & 0.41\\
		\textit{sim-building2}  & 76.05    				& 92.49  				& 21.60\%    &1.56          & 0.29\\
		\textit{sim-building3}  & 103.67     			& 140.02  				& 35.05\%    &1.62          & 0.41\\
		\textit{sim-building4}  & 84.05      			& 107.74  				& 28.18\%    &1.81          & 0.37\\
		\textit{real-building1} & 2667.33    			& 3025.60   			& 13.43\%    &2.06          & 0.66\\
		\textit{real-building2} & 4432.07    			& 4876.19 				& 10.02\%    &2.31          & 0.85\\
		\textit{real-building3} & 2495.45    			& 2896.52  				& 16.08\%    &4.70          & 0.72\\ 
		\hline
		
	\end{tabular}
\end{table*}

\subsection{Datasets}

To evaluate our entire pipeline, a cross-view fusion dataset is created which contains simulation data and real-world data. 
These datasets consist of aerial images captured by UAVs, LiDAR data, and ground-based video recordings. In Fig. \ref{fig:dataset}, a qualitative example of a dataset is shown.

Synthetic datasets are composed of over-view image set taken by UAV as well as  street-view video and LiDAR data taken by UGV (Unmanned Ground Vehicle). The environment is created using Unreal Engine 4, and Airsim simulation is used to control the UAV, UGV, and various sensors. For the over-view image set, the GNSS information is stored as well as the color image. For street-view scene, the camera and LiDAR are loaded in front of the UGV to capture the environmental information in real time and record the information through the recording function of ROS.

In order to verify the results of the algorithm, we enter the dataset in the actual world. The DJI Mavic 2 Pro UAV is used to cruise the urban area at a certain image repetition rate, take photos at regular intervals, record the GNSS position information when taking photos, and generate an over-view image set. The flight altitude is about 65 meters. For the collection of street-view environment on the ground side, we use RealSense D435 camera, Livox Avia LiDAR and GNSS receiver, which are connected to Jetson Xavier NX for data collection.

\subsection{Experiment Results and Analysis}

As shown in the Figs. \ref{fig:sim_ex} and \ref{fig:real_ex}, experiments are conducted on both simulation and real datasets. It compares the fusion data obtained through the GICP\cite{segal2009generalized} method and our method. The GICP algorithm is used with default parameters. The figure shows that ICP-based methods are not well-suited for supplementing incomplete point clouds, since they rely on an iterative point-to-point optimization process. Gaps in incomplete point clouds can cause the ICP loss function to rapidly increase, which forces the algorithm to search for matches in the gaps and can result in deviations in global registration. In contrast, our algorithm focuses on matching point cloud patches and prioritizes patches with prominent features for matching. Through an iterative coarse-to-fine registration process, the final matching error is reduced to tens of centimeters. Quantitative results are listed in Table \ref{table:evaluation_metrics}. It can be seen that the registration accuracy and filling effect are better than traditional methods.


\section{Conclusion}

In this paper, we propose a framework that can register cross-source point clouds and fill in missing points from various viewing angles. This framework has the following advantages:
1) High accuracy, our method can correctly fit cross-source point clouds, even if there is a large range of point missing. 
2) Point cloud filling and fusion: 3D cross-source point clouds reconstructed from a single viewing angle usually contain missing points. Through cross-source point cloud filling from multiple viewing angle, the gap of point clouds can be filled.

Although the proposed method achieves good performance, there are still some limitations. In noisy scenes, the registration effect may not be good enough. At this time, parameters are needed to adjust then get a better registration result.

\bibliographystyle{IEEEtran}
\bibliography{HybridFusion}

\end{document}